\documentclass[letterpaper, 10 pt, conference]{ieeeconf}  

\IEEEoverridecommandlockouts                              

\usepackage{amsmath} 
\usepackage{amssymb}  

\usepackage{graphicx}
\usepackage{float}
\usepackage{capt-of}
\usepackage{etoolbox}
\usepackage{adjustbox}
\usepackage{hyperref}

\title{\LARGE \bf 
Beyond Pixels: Leveraging Geometry and Shape Cues for Online Multi-Object Tracking
}

\author{Sarthak Sharma$^{1}$$^{*}$, Junaid Ahmed Ansari$^{1}$$^{*}$, J. Krishna Murthy$^{2}$, K. Mahdava Krishna$^{1}$ \\
$^{1}$ \small Robotics Research Center, KCIS, IIIT Hyderabad, India \\
$^{2}$ \small Mila, Universite de Montreal, Canada \\
$^{*}$ \small denotes equal contribution}

\begin{document}



\twocolumn[{%
\renewcommand\twocolumn[1][]{#1}%
\maketitle
\begin{center}
\centering
\includegraphics[width=\textwidth,height=5cm]{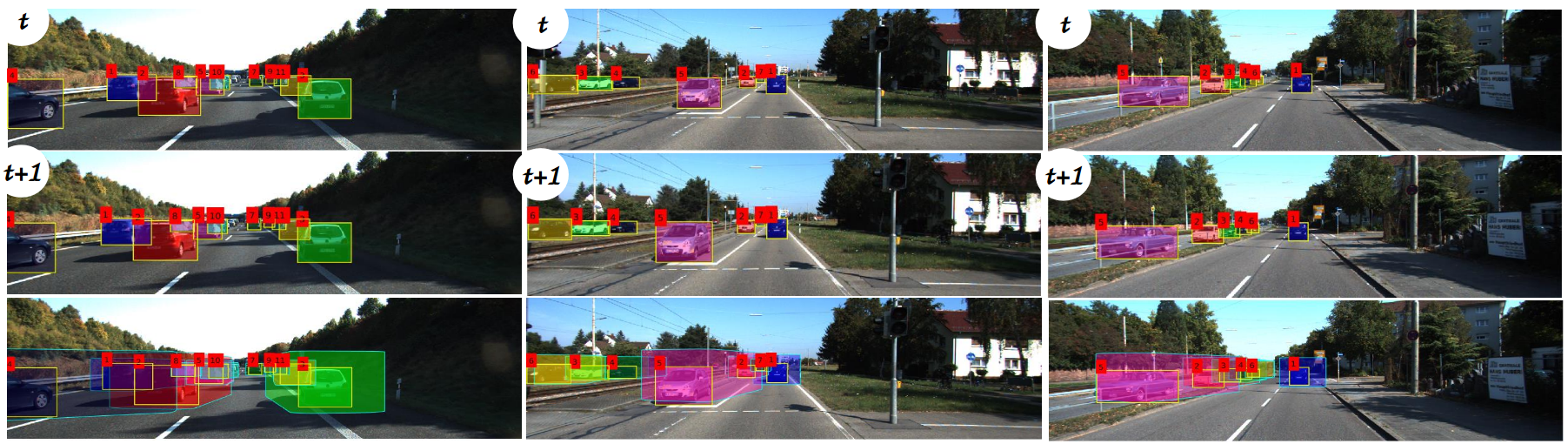}
\captionof{figure}{An illustration of the proposed method. The first two rows show objects tracks in frames $t$ and $t+1$. The bottom row depicts how 3D position and orientation information is propagated from frame $t$ to frame $t+1$. This information is used to specify search areas for each object in the subsequent frame, and this greatly reduces the number of pairwise costs that are to be computed.}
\label{fig:teaser}
\end{center}%
}]


\begin{abstract}
This paper introduces geometry and novel object shape and pose costs for multi-object tracking in road scenes. Using images from a monocular camera alone, we devise pairwise costs for object tracks, based on several 3D cues such as object pose, shape, and motion. The proposed costs are agnostic to the data association method and can be incorporated into any optimization framework to output the pairwise data associations. These costs are easy to implement, can be computed in real-time, and complement each other to account for possible errors in a tracking-by-detection framework. We perform an extensive analysis of the designed costs and empirically demonstrate consistent improvement over the state-of-the-art under varying conditions that employ a range of object detectors, exhibit a variety in camera and object motions, and, more importantly, are not reliant on the choice of the  association framework. We also show that, by using the simplest of associations frameworks (two-frame Hungarian assignment), we surpass the state-of-the-art in multi-object-tracking on road scenes. More qualitative and quantitative results can be found at \url{https://junaidcs032.github.io/Geometry_ObjectShape_MOT/}. Code and data to reproduce our experiments and results are now available at \url{https://github.com/JunaidCS032/MOTBeyondPixels}. 

\end{abstract}

\section{INTRODUCTION}

Object tracking in road scenes is an important component of urban scene understanding. With the advent and subsequent surge in autonomous driving technologies, accurate multi-object trackers are desirable in several tasks such as navigation and planning, localization, and traffic behavior analysis.

In this paper, we focus on designing a simple and fast, yet accurate and robust solution to the Multi-Object Tracking (MOT) problem in an urban road scenario. The dominant approach to multi-object tracking is tracking-by-detection, where the entire process is divided into two phases. The first phase comprises object detection, where bounding-boxes of objects of interests are obtained in each frame of the video sequence. The second phase is the data association phase, which is often the hardest step in the tracking-by-detection paradigm. Several factors such as spurious or missing detections, repeat detections, or occlusions and target interactions are confounding factors in this data association phase. 

Although several approaches \cite{DeepNetworkFlow,eccv2016,ijcv2017,greedyTracker,survey} exist for accurate online tracking of moving vehicles from a moving camera, most of them \cite{NOMT,eccv2016} use handcrafted cost functions that are either based on primitive features such as bounding box position in the image and color histograms, or are highly sophisticated and non-intuitive in design (eg. ALFD \cite{NOMT}). On the other hand, we propose costs that are intuitive, easy to compute and implement, and provide complementary cues about the target.

We exploit the fact that road scenes have a unique geometry and use this prior information to design costs. The proposed costs capture 3D cues arising from this scene geometry, as well as appearance based information. Further, we introduce a novel cost that captures similarity of 3D shapes and poses of target hypotheses. To this end we leverage recent work on shape-priors for object detection and localization from monocular sequences \cite{KM_ICRA,KM_IROS}. To the best of our knowledge, such pairwise costs have not been incorporated in multi-object tracking frameworks.

The efficacy of the monocular 3D cues is best portrayed in Fig.\ref{fig:teaser}. In this figure the first two rows illustrate the objects with their bounding boxes in two successive frames at $t$ and $t+1$. Upon lifting the objects at $t$ to 3D and ballooning their locations to account for large uncertainties in ego motion, we project them into the image observed at $t+1$. This gated/overlapping area shown in their respective colors in the last row of Fig.\ref{fig:teaser} reduces the search area for each such object significantly thereby reducing the pairwise costs. By backprojecting that lie only within this gated area into 3D and ascertaining data association costs based on 3D volume overlaps significantly improves tracking accuracy even with a straight forward Hungarian data association scheme.

The proposed costs are not too dependent on the choice of data association framework. We demonstrate the superiority of the proposed costs over monocular video sequences of urban road scenes that capture a wide range of camera and target motions, and also consistent improvement over other costs regardless of the choice of the object detector. We perform an extensive evaluation of various modes of the proposed costs on the KITTI Tracking benchmark \cite{KITTI} and obtain state-of-the-art performance, beating previous approaches by using a simple two-frame Hungarian association scheme. The approach is tested on KITTI online evaluation sever and outperforms the previous published approaches significantly. Naturally, more complex data association schemes, such as network flow based algorithms \cite{NetFlow2008,DiscreteContinuous,gmmcp,followme} can result in much better performance boosts upon incorporation of the proposed pairwise costs.

The paper contributes as follows.
\begin{enumerate}
\item It introduces novel data association cues based on single view reconstruction of objects that results in best tracking performance reported thus far in KITTI training datasets. It outperforms the nearest reported values in training data \cite{choi2013,DeepNetworkFlow,NOMT}, by at-least 12\% . The approach is tested on the KITTI Tracking online evaluation server where it outperforms the published approaches by a margin of over $6\%$.

\item It showcases that such improvements are sufficiently detector agnostic and repeatable over baseline appearance tracking based on object detectors such as \cite{RRC,SubCNN} through ablation studies

\item Finally it also identifies a role for 3D pose and shape cues where they play a role in improving tracking performance. 

\end{enumerate}

Monocular 3D cues especially based on single view geometry can often be unreliable. However when computed effectively they can be used reliably and repeatably even in challenging sequences such as KITTI. This constitutes the central theme of this effort.

\section{Related Work}

In this section, we review relevant work on multi-object tracking, and compare and contrast it with the proposed approach.

\subsection{Global Tracking}

Many approaches to tackle the association problem are \emph{global} \cite{NetFlow2008,gmmcp,everybody,DiscreteContinuous,KSSP,Chari}, in the sense that they assume detections from all frames are available for processing. Most global methods operate by mapping the tracking problem to a min-cost network flow problem. The original idea was proposed in \cite{NetFlow2008} and also provides for a method for explicit occlusion reasoning. An efficient variant is an approach based on generalized minimum clique graphs \cite{gmmcp}, where associations are solved for one object at a time while other objects are implicitly incorporated. Another section of global methods attempts to construct small chunks of trajectories (called tracklets), and compose them hierarchically to form longer trajectories, rather than solving for a min-cost flow over a densely connected graph.

\subsection{Online MOT}

In contrast to this, online trackers \cite{greedyTracker,gool2009,ijcv2017,CIWT} do not assume any knowledge of future frames and operate greedily, only with the data available upto the current instant. Such trackers often formulate the association problem as that of bipartite matching, and solve it via the Hungarian algorithm. A recent variant proposes near-online trackers \cite{NOMT}, in an attempt to provide the best of both worlds, i.e., to combine the capability of global methods to handle long-term occlusions and still achieve very low output latencies. Gieger et al \cite{followme} propose a memory and computation cost bound variant of network flow using dynamic programming.

Both these paradigms rely on handcrafted pairwise costs being fed into the association framework. Most of these are sophisticated in design and do not end up capturing 3D information that is easily available in road scenes.

\subsection{Learning Costs for MOT}

Significant attention has also been devoted to the task of learning pairwise costs for target tracking problems. In \cite{ijcv2017}, a structured SVM was used to learn pairwise costs for a bipartite matching data association framework. Other works have used graphical models, divide and conquer strategies and also learn unary costs. A more recent work \cite{DeepNetworkFlow} learns all costs using a deep neural network. On the other hand, we show that our simple, yet clean and efficient cost function designs significantly improve performance without the need of extensive hyperparameter search or cost learning.

\section{Problem Formulation}

We adopt the tracking-by-detection paradigm where we assume that we are provided with a monocular video sequence of $F$ frames $\{I_f\}$ for $f \in \{1 .. F\}$, and a set of object detections $D_f$ for each frame $I_f$. Each detection set consists of object detections $\{D^i_f\}$, where $i \in \{1 .. N_f\}$ ($N_f$ is the number of detections in frame $f$). Note that $D_f$ can also be an empty set, in the case where no objects are detected in a frame. Each detection $D^i_f$ is parametrized as $D^i_f = (x^i_f, y^i_f, w^i_f, h^i_f, s^i_f)$, where $(x^i_f,y^i_f)$ corresponds to the top-left corner of the detection box in the image, $w^i_f$ is the bounding box width, $h^i_f$ is the bounding box height, and $s^i_f$ is the detectors confidence in the bounding box (greater value indicates higher confidence). The multi-object tracking problem is to associate each bounding box to a target trajectory $T_k$ such that the following constraints are met. 
\begin{itemize}
\item Each target trajectory $T_k$ comprises of a set of bounding boxes (all from different frames) belonging to a unique target in the scene.
\item There are exactly as many trajectories $K$ as there are targets to be tracked.
\item In all frames where a target is visible, it is detected and assigned to the corresponding unique trajectory for the object.
\item All spurious bounding box detections are unassigned to any target trajectory.
\end{itemize}

The tracking problem formulated above is usually solved in a min-cost network flow framework (global tracking), a moving window dynamic programming framework (near-online tracking) or a bipartite matching framework (online tracking). Note that these are not the only available frameworks, but a representative set of most tracking approaches. All these frameworks (and the others not mentioned here) use pairwise costs to define affinity across pairs of detections. The association framework then computes a Maximum A Posteriori (MAP) estimate of the target trajectory $T_k$, given the detection hypotheses $D = \left(D_1, D_2, ... D_f\right)$ and an affinity matrix that gives the likelihood of each detection in each frame corresponding to every detection in every other frame.

\section{Geometry and Object Shape Costs}

The core contribution of this paper is to design intuitive pairwise costs that are efficient to compute, and are accurate for tracking. We focus on urban driving scenarios and demonstrate how the geometry of urban road scenes can be exploited to infer 3D cues for tracking.

Typical costs in tracking algorithms include bounding box locations, trajectory priors, optical flow, bounding box overlap, and appearance information (color histograms or path-based cross-correlation measures). These costs require careful handcrafting, finetuning, and hyperparameter estimation. We propose to use a set of simple complementary costs that are readily available from recent monocular 3D object localization systems \cite{KM_ICRA,KM_IROS}. We also introduce a novel cost based on the 3D shape and pose of the target. We show that this cost, apart from improving data association performance, also assists in discarding false detections without incurring large computational overhead.

\subsection{System Setup}

We focus on autonomous driving scenarios, where the video sequence is from a monocular camera mounted on a car moving on the road plane, and the targets to be tracked are also moving on the road. Feature based odometry is run on a background thread (for rough frame-to-frame motion estimation). Also, we make use of a recent approach that goes beyond bounding boxes and estimates the 3D shape and pose of objects, given just a single image \cite{KM_ICRA}. This is done by lifting discriminative parts in 2D (\emph{keypoints}) to 3D. These keypoints are a set of points chosen so that they are common across all object instances (eg. for a car, we have centers of wheels, headlights, taillights, etc). The authors use a CNN architecture \cite{KM_IROS} to localize these keypoints in 2D, given a detection.

The 3D shape of the object is parametrized as the sum of the mean shape (for the object category) and a linear combination of so-called \emph{basis shapes}. Mathematically, 
\begin{equation}
S = \bar{S} + \sum_{b=1}^b \lambda_b V_b
\end{equation}
where $S$ is the shape of a particular instance, $\bar{S}$ is the mean shape for the object category, and $V_b$ is the deformation basis (a set of eigenvectors) that characterizes deformation directions of the mean shape. We use the same model in \cite{KM_ICRA} and denote the shape vector of an object by $\Lambda = \left[\lambda_1 .. \lambda_B\right]^T$, where $B$ is the number of vectors in the deformation basis (typically, $B = 5$). 

The pipeline in \cite{KM_ICRA} also estimates the 3D pose of the object, which is parametrized as an axis-angle vector $\omega$. Moreover, an estimate of object dimensions (height, width, and length) is also returned.

\subsection{3D-2D Cost}

\begin{figure}[!t]
  \centering
  \includegraphics[width=0.5\textwidth]{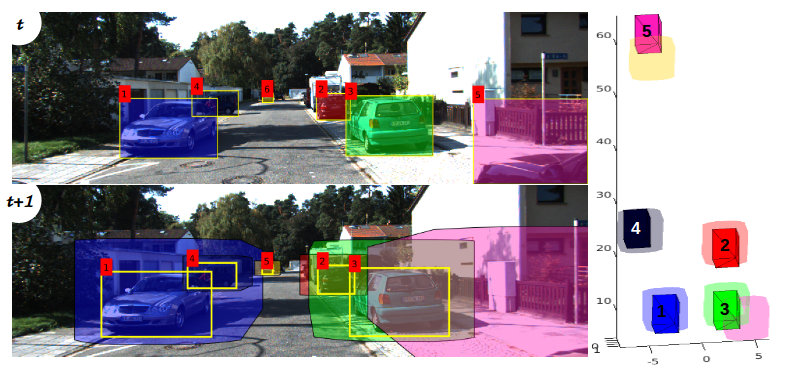}
  \caption{Illustrating the concept of 3D-2D and 3D-3D costs Two subsequent frames, t and t+1, are shown in the left. For each detection in the frame t, we compute and propagate (with uncertainty) it’s 3D bounding box in the next frame t + 1. These boxes are projected to 2D in the frame t+1. The intersection between the detection boxes of t+1 and these projections constitutes the 3D-2D cost. The intersection of the 3D bounding boxes in 3D constitute the 3D-3D cost as shown in the right; propagated bounding boxes are colored with their respective 2D box in frame t and 3D bounding boxes of  detections in frame t+1 are numbered respectively.}
  \label{fig:3d2d}
\end{figure}

Given the height $h_{cam}$ of the camera above the ground, assuming that the bottom line of each bounding box detection $d^i_f$ in frame $f$ is on the road plane, a depth estimate of the car in the current camera coordinates can be obtained by back projection via the road plane as in \cite{chandraker2015}, using 
\begin{equation}
X_f = \pi_G^{-1}(x) = \frac{h\mathbf{K}^{-1}x}{n^T\mathbf{K}^{-1}x}
\label{eqn:backprojection}
\end{equation}
where $x$ is the bottom center of the detected bounding box, \textbf{K} is the camera intrinsic matrix and $\pi_G^{-1}$ is used as shorthand for backprojection via the ground plane. This backprojection equation is only accurate when $x$ is known precisely, which is not usually the case. Hence, we estimate the uncertainty in 3D location of $X_f$ by using a linearized version of (\ref{eqn:backprojection}) and assuming that the detector confidence is an isotropic 2D Gaussian, i.e., $(x^i_f, y^i_f)^T \sim \mathcal{N}(0,\sigma^2\mathbf{I}_{2\times2})$. This region is expanded (anisotropically) by the estimates of the target dimensions returned by the system \cite{KM_ICRA}.

Now, assume we have another detection $d^j_{f'}$ in frame $f'$ with which we wish to compute the pairwise affinity of $d^i_f$. We obtain a rough estimate of the camera motion from frame $f$ to frame $f'$ using a feature-based odometry thread running in the background. Using this estimate of the camera motion, we transport $X_f$ to the camera coordinates of frame $f'$, while duly accounting for the uncertainty in camera motion estimate, and in the backprojection via the road plane. The obtained coordinates $X_{f'}$ are then projected down to the image frame $f'$ to obtain a 2D search area in which potential matches for $X_f$ are expected to be found, as shown in the  frame $t+1$ of Fig.\ref{fig:3d2d}. Mathematically, the 3D-2D cost for two detections $d^i_f$ and $d^j_{f'}$ is defined as follows
\begin{equation}
\mathcal{C}_{3D2D}(d^i_f, d^j_{f'}) = 1 - \frac{(\pi(g(\xi, \phi(\pi_G^{-1}(y^i_f), s^i_f)) \cap b^j_{f'}))}{b^j_{f'}}
\label{eqn:3D-2D}
\end{equation}
Intuitively, this cost measures a (weighted) overlap of the 2D region in which the target is expected in frame $f'$ and the detection $d^j_{f'}$. $\pi$ denotes the projection operator that projects a 3D point to image pixel coordinates. $g(\xi,X)$ denotes a rigid-body motion $\xi \in se(3)$ applied to a 3D point $X \in \mathbb{R}^3$. $\phi(X,s)$ denotes the function that estimates the uncertainty of the 3D point $S$ according to a linearized form of (\ref{eqn:backprojection}) and the detector confidence $s$.

Most importantly, this cost is evaluated only for detections $d^j_{f'}$ that lie inside the expected target area $(\pi(g(\xi, \phi(\pi_G^{-1}(y^i_f), s^i_f))$. This significantly reduces the number of comparisons needed to be made among target pairs.

\subsection{3D-3D cost}

Although useful in reducing the number of candidate detections to be evaluated, the 3D-2D cost has frequent confounding cases. This is because, we still measure overlap in the image space. To mitigate this drawback, we define a 3D-3D cost, which, instead of measuring 2D overlap, measures overlap in 3D,as shown in Fig.\ref{fig:3d2d} (right side). Here, we backproject each candidate $d^j_{f'}$ via the road plane, and measure overlap with respect to the transformed 3D volume from frame $f$ given by $g(\xi, \phi(\pi_G^{-1}(y^i_f), s^i_f)$. The 3D-3D cost for two detections $d^i_f$ and $d^j_{f'}$ is defined as
\begin{equation}
\mathcal{C}_{3D3D}(d^i_f, d^j_{f'}) = 1 - \frac{g(\xi, \phi(\pi_G^{-1}(y^i_f), s^i_f))}{\phi(\pi_G^{-1}(b^j_{f'}),s^j_{f'})}
\label{eqn:3D-3D}
\end{equation}

In order to speed up evaluation of 3D overlap, we exploit the inherent geometry of road scenes. Since all objects of interest are on the road plane (the XZ plane in our case), it is sufficient to measure overlap in the XZ plane. This is because all objects are at nearly constant heights above the ground and hence have similar overlap in the Y direction.

\subsection{Appearance Cost}
\begin{figure}
  \centering
  \includegraphics[width=0.5\textwidth]{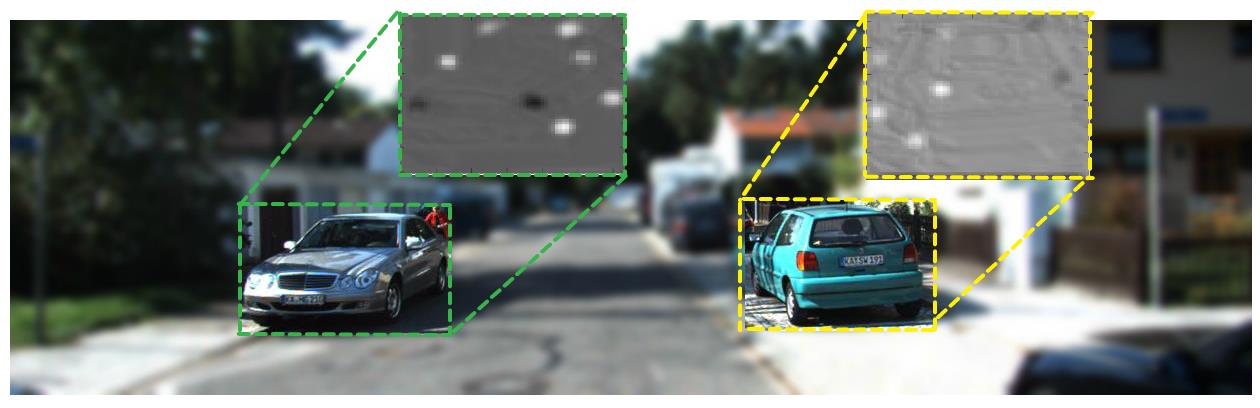}
  \caption{Weighted combination of the features captured by the hourglass network. Such a descriptor is able to capture the dissimilarity between different detections.}
  \label{fig:2d2d}
\end{figure}
In \cite{KM_IROS}, the authors train a stacked-hourglass CNN architecture to localize a discriminative set of keypoints on an image. This deep CNN architecture captures various discriminative features for each detection, along with the keypoint evidence. We use weighted combination of activation maps from the output of the layers of the hourglass network as a feature descriptor for each detection, as shown in Fig.\ref{fig:2d2d}  and compute a similarity score between detections using the L$2$ Norm between descriptors from the image patch inside each of the bounding boxes. If $\psi(.)$ denotes the feature descriptor of each detection, the appearance cost is defined as
\begin{equation}
\mathcal{C}_{app} (d^i_f, d^j_{f'}) = \eta_{app} \| \psi(d^i_f) - \psi(d^j_{f'})  \|_2^2
\label{eqn:appearance}
\end{equation}
where $\eta_{app}$ is a normalization constant.

\subsection{Shape and Pose Cost}

We use a novel shape and pose cost based on the single image shape and pose returned by the pipeline of \cite{KM_ICRA} . Shape is parameterized as a vector comprising of deformation coefficients $\Lambda = \left[\lambda_1 .. \lambda_B\right]^T$, where $B$ is the number of deformation basis vectors (usually $5$). Each possible value of $\Lambda$ denotes a unique class of object instances and hence carries useful information about the 3D shape of the target. For instance varying certain parameters of $\Lambda$ may represent a shape that is more SUV-like than Sedan-like, and so on. Pose is parametrized as an axis-angle vector $\omega$. For detections $d^i_f$ and $d^j_{f'}$, the shape and pose cost is specified as
\begin{equation}
\mathcal{C}_{s} (d^i_f, d^j_{f'}) = \eta_{s} \| \Lambda(d^i_f) - \Lambda(d^j_{f'})  \|_2^2  +  \eta_{p} \| \omega(d^i_f) - \omega(d^j_{f'})  \|_2^2
\label{eqn:shape}
\end{equation}
where $\eta_{s}$ and $\eta_{p}$ are normalization constants.

The overall pairwise cost term is a weighted linear combination of all the aforementioned cost. The weights of the linear combination are determined by four-fold cross validation on the train set.

\section{Results}

In this section, we present an account of the experiments we performed, and we report and analyze the findings thereof. In nutshell, we evaluate our tracking framework on a variety of challenging urban driving sequences and demonstrate a substantial performance boost over the state-of-the-art in multi-object tracking, by using the simplest of tracking frameworks, viz. bipartite matching using the Hungarian algorithm.

\subsection{Dataset}
We evaluate the proposed multi-object tracking framework on the popular KITTI Tracking benchmark on both training as well as testing dataset. \cite{KITTI}. As prescribed in \cite{DeepNetworkFlow,NOMT,KITTI}, we divide the training dataset, which contains 21 sequences, into four splits,for cross validation. The splits are chosen so that each split contains a similar distribution of number of vehicles per sequence, occlusion and truncation levels, and relative motion patterns between the camera and the target. The cross validation helps us to tune the weight for each of the proposed costs to compute the final cost matrix. The best performing combination of these weighted costs are used for reporting the result on the KITTI Tracking benchmark. Multiple vehicles moving with varying speeds, variance in the ego camera motion, and target objects appearing in non conforming locations in frames make the KITTI Tracking dataset \cite{KITTI} a truly challenging one. We report results on the \emph{Car} class.

\begin{table*}[!hbt]
	\centering
    
    \caption{Results on the KITTI Tracking train set. Tracking Accuracy(MOTA) and Precision(MOTP), Mostly Tracked(MT),Partly Tracked(PT), Mostly Lost(ML) , True Positives(TP) , False Positives(FP), ID-switches(IDS), Fragmentation(FRAG) }
    
    \begin{adjustbox}{max width=\linewidth}    
	\begin{tabular}{|c|c|c|c|c|c|c|c|c|c|c|c|}
		\hline\hline
    	 &  MOTA & MOTP & Recall & Precision & MT & PT & ML & TP & FP & IDS & FRAG \\
      	\hline\hline
     	CIWT \cite{CIWT} & $74.38$ & $82.85$ & - & - & $49.59$ & $40.68$ & $9.80$ & - & - & $\textbf{26}$ & $\textbf{131}$ \\
    	\hline
        Ours (On split of \cite{CIWT}) & $\textbf{91.75}$ & $\textbf{89.90}$ & $94.83$ & $98.62$ & $\textbf{88.61}$ & $10.39$ & $\textbf{0.9}$ & $9814$ & $137$ & $93$ & $151$ \\
    	\hline\hline
    	Deep Network Flow \cite{DeepNetworkFlow} & $74.11$ & - & $84.74$ & $92.05$ & $61.73$ & - & - & - & - & $\textbf{29}$ & $\textbf{335}$ \\
        \hline
        NOMT \cite{NOMT} & $73.07$ & - & $85.07$ & $90.92$ & $61.73$ & - & - & - & - & $43$ & $386$ \\
        \hline
        SSP \cite{followme} & $67$ & $79$ & - & - & $41$ & - & $9$ & - & - & 194 & 977 \\
    	\hline
    	Ours & $\textbf{91.4}$ & $\textbf{89.83}$ & $\textbf{94.65}$ & $\textbf{98.47}$ & $\textbf{87.76}$ & $10.63$ & $\textbf{1.59}$ & $25309$ & $392$ & $232$ & $423$ \\
    	\hline
        
	\end{tabular}
    \end{adjustbox}
    \label{table:overall}
\end{table*}

\begin{table*}[!hbt]
	\centering
    
    \caption{Results on the KITTI Tracking test set. For more details, visit \url{http://www.cvlibs.net/datasets/kitti/eval_tracking.php}}
    
    \begin{adjustbox}{max width=\linewidth}    
	\begin{tabular}{|c|c|c|c|c|c|c|}
		\hline\hline
    	 &  MOTA & MOTP & MT & ML & IDS & FRAG \\
      	\hline\hline
        CIWT \cite{CIWT} & $75.39$ & $79.25$ &$49.85$ & $10.31$ & $165$ & $660$  \\
        \hline
        SCEA \cite{SCEA} & $75.58$ & $79.39$ &$53.08$ & $11.54$ & $104$ & $448$  \\
        \hline
     	MDP \cite{MDPTracker} & $76.59$ & $82.10$ &$52.15$ & $13.38$ & $130$ & $387$   \\
    	\hline
        LP-SSVM \cite{ijcv2017} & $77.63$ & $77.80$ &$56.31$ & $8.46$ & $62$ & $539$  \\
        \hline
       NOMT \cite{NOMT} & $78.15 $ & $79.46$ &$57.23$ & $13.23$ & $\textbf{31}$ & $\textbf{207}$  \\
        \hline
       MCMOT-CPD \cite{eccv2016} & $78.90$ & $82.13 $ &$52.31$ & $11.69$ & $228$ & $536$  \\
    	\hline
    	Ours(RRC-IIITH) & $\textbf{84.24}$ & $\textbf{85.73}$ &$\textbf{73.23}$ & $\textbf{2.77}$ & $468$ & $944$  \\
    	\hline
        
	\end{tabular}
    \end{adjustbox}
    \label{table:test}
\end{table*}

\begin{table*}[!hbt]
	\centering
    \caption{Ablation Study. Comparision across various cues used for pairwise cost computation and choice of object detector. (App - appearance cost)}
    \begin{adjustbox}{max width=\linewidth}
	\begin{tabular}{|c|c|c|c|c|c|c|c|c|c|c|c|c|c|c|}
		\hline 
        Cue(s) &  MOTA & MOTP & Recall & Precision & MT & PT & ML & TP & FP & IDS & FRAG \\
      	\hline\hline
        
		SubCNN + App & $66.18$ & $82.60$ & $86.45$ & $89.29$ & $71.63$ & $24.46$ & $3.90$ & $23563$ & $2825$ & $708$ & $1100$ \\
            \hline
		SubCNN + 3D-2D & $68.24$ & $82.60$ & $86.45$ & $89.29$ & $71.63$ & $24.46$ & $3.90$ & $23563$ & $2825$ & $429$ & $829$ \\
             	\hline
		SubCNN + 3D-3D & $69.36$ & $82.60$ & $86.45$ & $89.29$ & $71.63$ & $24.46$ & $3.90$ & $23563$ & $2825$ & $377$ & $778$ \\
      \hline
		SubCNN + 3D-2D + App & $70.96$ & $82.60$ & $86.45$ & $89.29$ & $71.63$ & $24.46$ & $3.90$ & $23563$ & $2825$ & $472$ & $870$ \\
    	\hline
		SubCNN + 3D-3D + 3D-2D + App + Shape-Pose & $\textbf{71.52}$ & $\textbf{82.60}$ & $\textbf{86.45}$ & $\textbf{89.29}$ & $\textbf{71.63}$ & $\textbf{24.46}$ & $\textbf{3.90}$ & $\textbf{23563}$ & $\textbf{2825}$ & $\textbf{338}$ & $\textbf{740}$ \\
        
        \hline\hline
    	
		RRC + App & $80.53$ & $89.83$ & $94.62$ & $98.47$ & $87.76$ & $10.63$ & $1.59$ & $25309$ & $392$ & $2863$ & $3022$ \\
            \hline
		RRC + 3D-2D & $86.91$ & $89.83$ & $94.65$ & $98.47$ & $87.76$ & $10.63$ & $1.59$ & $25309$ & $392$ & $1328$ & $1507$ \\
             	\hline
		RRC + 3D-3D & $87.56$ & $89.83$ & $94.65$ & $98.47$ & $87.76$ & $10.63$ & $1.59$ & $25309$ & $392$ & $1170$ & $1333$ \\
      \hline
		RRC + 3D-2D + App & $89.65$ & $89.83$ & $94.65$ & $98.47$ & $87.76$ & $10.63$ & $1.59$ & $25309$ & $392$ & $668$ & $849$ \\
    	\hline
		RRC + 3D-3D + 3D-2D + App + Shape-Pose & $\textbf{91.4}$ & $\textbf{89.83}$ & $\textbf{94.65}$ & $\textbf{98.47}$ & $\textbf{87.76}$ & $\textbf{10.63}$ & $\textbf{1.59}$ & $\textbf{25309}$ & $\textbf{392}$ & $\textbf{232}$ & $\textbf{423}$ \\
        
        \hline
	\end{tabular}
    \end{adjustbox}
    \label{table:ablation}
\end{table*}


\begin{table*}[!hbt]
	\centering
    \caption{Results using Shape and Pose along with other Cues }
	\begin{tabular}{|c|c|c|c|c|c|c|c|c|c|c|c|}
		\hline\hline
    	 &  MOTA & MOTP & Recall & Precision & MT & PT & ML & TP & FP & IDS & FRAG \\
      	\hline\hline
		w/o Shape and Pose & $55.3$ & $86.01$ & $98.9648$ & $84.75$ & $100$ & $0$ & $0$ & $478$ & $86$ & $5$ & $7$  \\
        \hline
        with Shape and Pose & $\textbf{57.29}$ & $86.01$ & $98.9648$ & $84.75$ & $100$ & $0$ & $0$ & $478$ & $86$ & $\textbf{1}$ & $\textbf{5}$  \\
          
      \hline
 	\end{tabular}
    \label{table:shape_pose}
\end{table*}

\subsection{Evaluation Metrics}
To evaluate the performance of our approach, we adopt the widely used CLEAR MOT metrics \cite{CLEARMOT}. The overall performance of the tracker is summed up in two intuitive metrics, viz. Multi-Object Tracking Accuracy (MOTA) and Multi-Object Tracking Precision (MOTP). While MOTA is concerned with tracking accuracy, MOTP deals with object localization precision.

\subsection{System Overview}
The proposed approach is a tracking-by-detection approach and hence assumes per-frame bounding box detections as input. We choose two recent object detectors --- Recurrent Rolling Convolution (RRC) \cite{RRC} and SubCNN \cite{SubCNN}. Each of these detectors provides multiple detections per frame. A threshold is applied on the detection scores and those detections whose confidence scores are lower than the threshold are pruned. In addition to this, we run a non-maxima suppression (NMS) scheme to subdue multiple detections around the same object. These detections are used to compute pairwise costs as outlined in the previous section. These pairwise costs constitute a cost matrix that is used for a bipartite matching algorithm that associates detections across two frames. In practice, bipartite matching is performed using the $O(n^3)$ Hungarian algorithm \cite{hungarian}.

\subsection{Approaches Considered}
We evaluate the proposed framework against the current best competitors on the KITTI Tracking dataset. We consider approaches \cite{DeepNetworkFlow,NOMT,CIWT,followme}. 

\subsection{Performance Evaluation}
We evaluate the performance of our approach on the current best competitors on the KITTI Tracking Benchmark. While \cite{NOMT,CIWT,followme} rely on complex handcrafted costs, \cite{DeepNetworkFlow} learns all unary and pairwise costs that are input to a network flow based tracker. Moreover, the data association steps of \cite{NOMT,CIWT,followme} rely on complex optimization routines. The proposed approach is also evaluated on the KITTI Tracking evaluation sever.

Table \ref{table:overall},where we compare our two-frame based approach with the other competitors using the best performing object detector in the form of \cite{RRC} and a judicious combination of such appearance, 3D, pose and shape cues best possible results on KITTI training sequence are achieved in terms of MOTA ($91.4\%$) and MOTP ($89.84\%$). Although our method suffers from ID switches and fragmentations, this is typical of online trackers; more so of two-frame greedy trackers. Using the proposed pairwise costs in a slightly more sophisticated tracker such as \cite{NOMT,followme} will naturally reduce ID switches and fragmentations also.

Table \ref{table:test},where we compare our two-frame based approach with the other published approaches on the KITTI Tracking online server. We outperform the next best competitor by a margin of ($6\%$) on the test set, achieving state of the art results in the form of MOTA ($84.24\%$), MOTP ($85.73\%$), MT ($73.23\%$) and ML ($2.77\%$).

\subsection{Ablation Study}

We then perform a thorough ablation analysis of various cues used for computing pairwise costs across two distinct object detectors: RRC \cite{RRC} and SubCNN \cite{SubCNN}. 
Results are summarized in Table \ref{table:ablation}. This analysis captures the importance of each of the proposed cue and demonstrates that the combination of all these is crucial for overall performance. Notice how each cue improves the performance of our system in terms of MOTA ,ID switches and fragmentations. Even with underperforming detectors such as \cite{SubCNN}, there is a tangible performance boost by using a combination of monocular 3D cues. This is portrayed in ablation analysis of SubCNN detectors in Table \ref{table:ablation}. Furthermore the repeatability of performance gain using these novel cues over any baseline detection methods is also delineated.

There exist subsequences where the role played by shape and pose cues become relevant. While in a typical road scene involving lane driving the pose cues are not discriminatory (as the vehicles are aligned with the lane direction), they become discerning enough in areas such as intersections, round abouts where pose and viewpoint changes are heterogeneous. This is showcase in Table \ref{table:shape_pose}. Here, we select particular frames from the KITTI Tracking dataset, which  have images containing cars moving at intersections, which captures different viewpoints and shapes of cars. Using detections from a weak detector \cite{SubCNN} and a simplistic combination 2D-2D cues along with shape and pose cue of the car performs better than the stand alone 2D cue, for sequences which have cars with various viewpoints over the frames.

\subsection{Qualitative Results}

\begin{figure*}[!t]
\centering
\includegraphics[width=\textwidth]{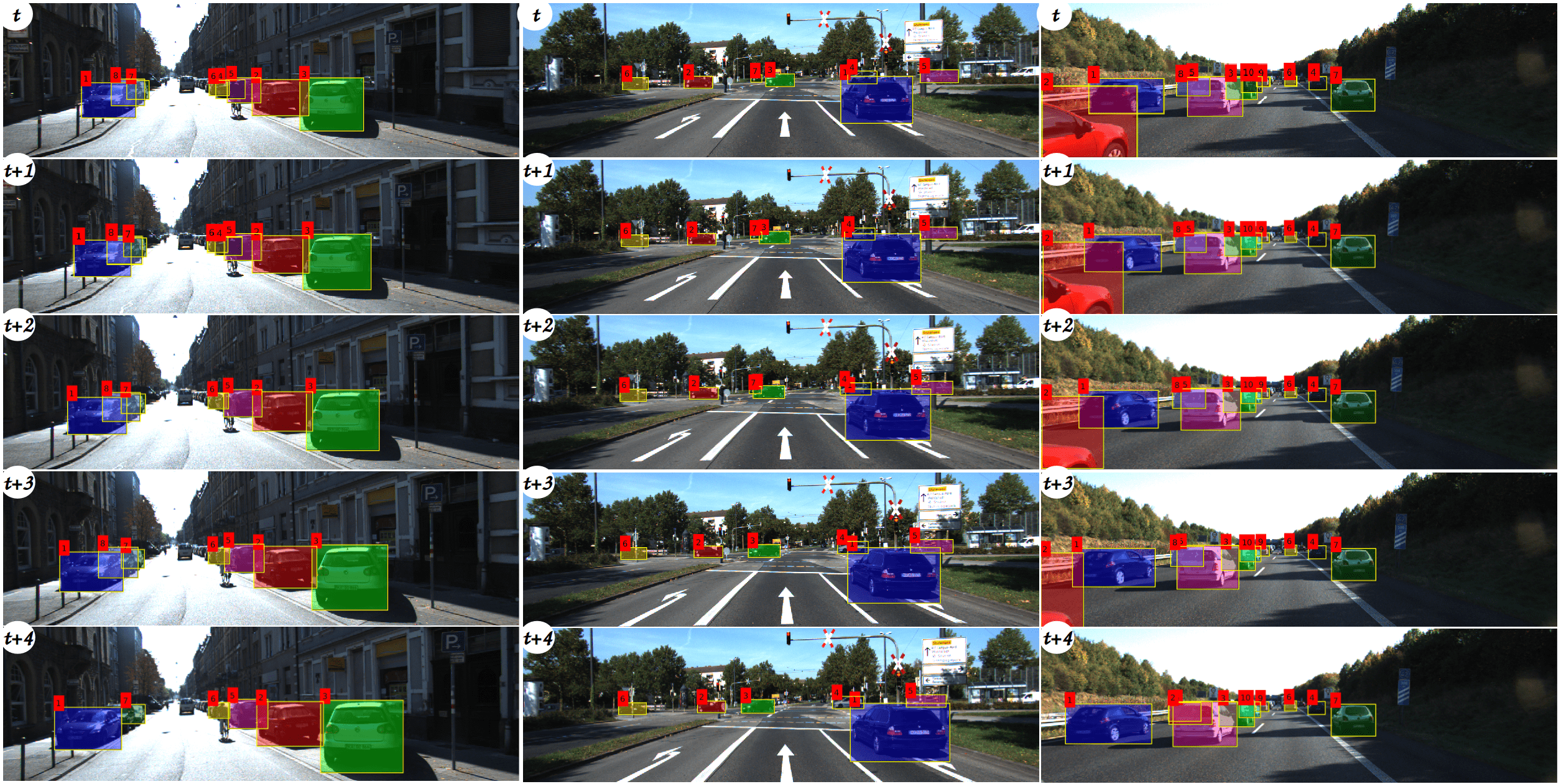}
\caption{Qualitative results on some challenging sequences.}
\label{fig:qualitative}
\end{figure*}

\begin{figure*}[!t]
\centering
\includegraphics[width=\textwidth]{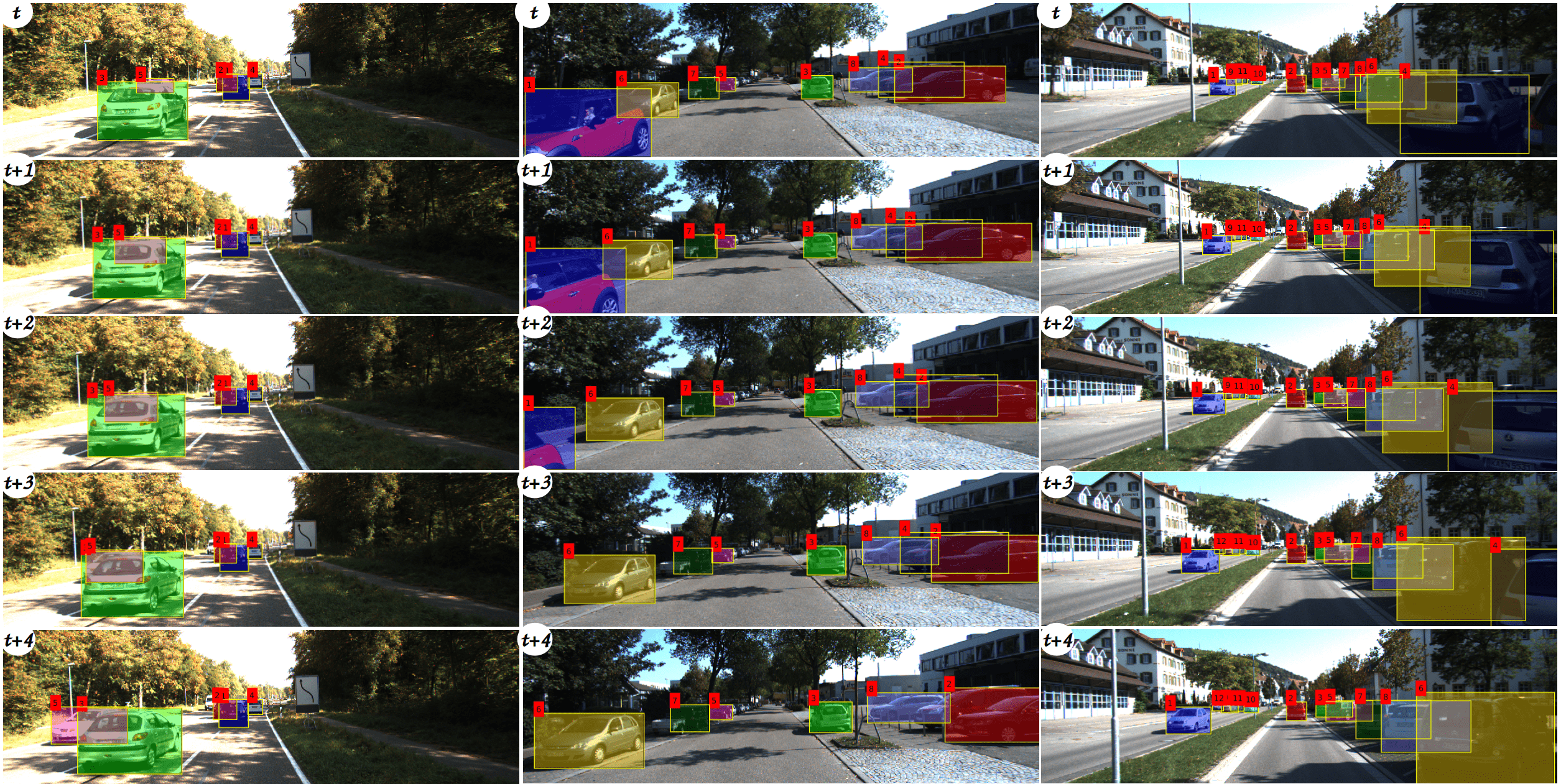}
\caption{Qualitative results on some challenging equences.}
\label{fig:qualitative2}
\end{figure*}

Finally, we present qualitative results from challenging sequences in Fig.\ref{fig:qualitative} and Fig.\ref{fig:qualitative2}. 
These results clearly indicate the ability of the proposed pairwise costs to disambiguate and track across viewpoint variations, clutter, and varying relative motion between the camera and the target.

For example the first column of Fig \ref{fig:qualitative} shows cars occluded on either sides of the road accurately tracked almost till the horizon. Whereas the second column shows efficient tracking of cars at varying depths and varying poses in an intersection while the third column shows precise tracking of occluding cars as well as a car that is being overtaken from the right by the ego car. In fact  in the 4th frame a very small portion of the car is visible yet accurately tracked.

\subsection{Summary of Results}

The cornerstone of this effort is that single view monocular 3D cues obtained though formalisms developed on the basis of single view geometry can be effectively exploited to track vehicles in challenging scenes. This gets illustrated in the various tabulations of this section. 

Table \ref{table:overall} depicts significant improvements over many of the current state of the art methods with a tracking accuracy in excess of $90\%$. 
We test our approach on the KITTI Tracking online server. Table \ref{table:test} depicts significant improvements over the published approaches, with tracking accuracy over $84\%$.

Whereas the ablation studies in Table \ref{table:ablation} does showcase the repeatability of 3D cues in improving the baseline appearance only tracking over detectors. While not as significant as in \cite{RRC} baseline improvement over SubNN object detector\cite{SubCNN} can be gleaned from Table \ref{table:ablation}. The improvement in ID switches and fragmentations can also be seen over both detector baselines as a consequence of the 3D cues. 

Table \ref{table:shape_pose} shows the relevance of pose and shape cues over a subsequence where association costs due to such cues improves baseline performance. 

\section{Conclusions}

Most state of the art tracking formalisms have not explored the role of 3D cues and when they have done those cues have been due to immediately available stereo depth. This paper showcased for the first time monocular 3D cues obtained from single view geometry along with pose and shape cues results in the best tracking performance on popular object tracking training datasets. These cues result in a set of simple, intuitive pairwise costs for multi-object tracking in a tracking-by-detection setting. Despite being more difficult to compute than ready made 3D depth data, monocular 3D cues have a role to play in diverse on road applications including object and vehicle tracking. Apart from the quantitative,  qualitative  results too signify its advantage in challenging scenes that involve considerable occlusions, minimal appearance of the object in the scene and objects that are far enough that they appear on the horizon. Although we demonstrated results using a simple Hungarian method based tracker, incorporation of sophisticated trackers would result in even higher performance boosts.


\bibliography{references}
\bibliographystyle{IEEEtran}

\end{document}